\documentclass[journal]{IEEEtran}
\usepackage{graphicx}
\usepackage{cite}
\usepackage{picinpar}
\usepackage{amsmath}
\usepackage{url}
\usepackage{flushend}
\usepackage{colortbl}
\usepackage{soul}
\usepackage{multirow}
\usepackage{pifont}
\usepackage{color}
\usepackage{alltt}
\usepackage[hidelinks]{hyperref}
\usepackage{enumerate}
\usepackage{siunitx}
\usepackage{epstopdf}
\usepackage{pbox}

\usepackage{amssymb}
\usepackage{stfloats}
\usepackage{float}

\usepackage{xcolor}
\usepackage[ruled,linesnumbered]{algorithm2e}
\usepackage{tabularx}
\usepackage{multirow}
\newcommand{\tabincell}[2]{\begin{tabular}{@{}#1@{}}#2\end{tabular}}

\DeclareSymbolFont{largesymbol}{OMX}{yhex}{m}{n}
\DeclareMathAccent{\Widehat}{\mathord}{largesymbol}{"62}

\ifCLASSINFOpdf
\else
\fi

\begin{document}
%

\title{On Reciprocally Rotating Magnetic Actuation of a Robotic Capsule in Unknown Tubular Environments}

%
%

\author{Yangxin~Xu$^{\star}$,~\IEEEmembership{Student~Member,~IEEE,}
        Keyu~Li$^{\star}$,~\IEEEmembership{Student~Member,~IEEE,}\\
        Ziqi~Zhao,
        and~Max~Q.-H.~Meng$^{\sharp}$,~\IEEEmembership{Fellow,~IEEE}
\thanks{This work is partially supported by National Key R \& D program of China with Grant No. 2019YFB1312400 and Hong Kong RGC CRF grant C4063-18G awarded to Max Q.-H. Meng.}
\thanks{Y. Xu and K. Li are with the Department of Electronic Engineering, the Chinese University of Hong Kong, Hong Kong SAR, China (e-mail: yxxu@link.cuhk.edu.hk; kyli@link.cuhk.edu.hk).}
\thanks{Z. Zhao is with the Department of Electronic and Electrical Engineering, the Southern University of Science and Technology, Shenzhen, China (e-mail: zzq2694@163.com).}
\thanks{Max Q.-H. Meng is with the Department of Electronic and Electrical Engineering of the Southern University of Science and Technology in Shenzhen, China, on leave from the Department of Electronic Engineering, the Chinese University of Hong Kong, Hong Kong SAR, China, and also with the Shenzhen Research Institute of the Chinese University of Hong Kong, Shenzhen, China (e-mail: max.meng@ieee.org).}

\thanks{$^{\star}$ The authors contribute equally to this paper.}
\thanks{$^{\sharp}$ Corresponding author.}
}

\maketitle

\begin{abstract}
Active wireless capsule endoscopy (WCE) based on simultaneous magnetic actuation and localization (SMAL) techniques holds great promise for improving diagnostic accuracy, reducing examination time and relieving operator burden. 
To date, the rotating magnetic actuation methods have been constrained to use a continuously rotating permanent magnet.
 In this paper, we first propose the reciprocally rotating magnetic actuation (RRMA) approach for active WCE to enhance patient safety. We first show how to generate a desired reciprocally rotating magnetic field for capsule actuation, and provide a theoretical analysis of the potential risk of causing volvulus due to the capsule motion. Then, an RRMA-based SMAL workflow is presented to automatically propel a capsule in an unknown tubular environment.
We validate the effectiveness of our method in real-world experiments to automatically propel a robotic capsule in an ex-vivo pig colon. 
The experiment results show that our approach can achieve efficient and robust propulsion of the capsule with an average moving speed of $2.48 mm/s$ in the pig colon, and demonstrate the potential of using RRMA to enhance patient safety, reduce the inspection time, and improve the clinical acceptance of this technology.
\end{abstract}

\begin{IEEEkeywords}
Medical robots and systems, Robot programming, Magnetic actuation and localization, Wireless capsule endoscopy.
\end{IEEEkeywords}

%
\IEEEpeerreviewmaketitle

\section{Introduction}

\IEEEPARstart{B}{ased} on the data reported in \cite{chan2019gastrointestinal}, about $8$ million people worldwide die from diseases in the gastrointestinal (GI) tract every year. The GI cancers have been among the top four leading causes of cancer death worldwide, accounting for $17.1\%$ of cancer deaths \cite{siegel2020colorectal}, and the higher incidence of GI cancers has increased the hospitalization burden \cite{chan2019gastrointestinal}. Optical endoscopy is considered the gold standard for screening the GI tract for early diagnosis and early treatment in clinical practice, which can significantly decrease the incidence and mortality of gastrointestinal cancers \cite{chen2021effectiveness}\cite{li2017time}. However, the manual screening process in standard optical endoscopy (e.g., colonoscopy) requires an experienced clinician to manually insert the endoscope and perform inspections, which would bring a heavy cognitive and physical burden on the clinician and may cause patient discomfort \cite{norton2019intelligent}. Moreover, the lack of medical resources for endoscopic screening in rural areas has resulted in the urban-rural disparity in GI cancer in some developing countries \cite{wen2017urban}.

Wireless Capsule Endoscopy (WCE) has become a promising new tool for painless and non-invasive inspection of the entire GI tract since 2000 \cite{iddan2000wireless}, but the whole examination process is very time-consuming, which usually takes about $8 \sim 24$ hours, as the capsule is passively actuated by the intestinal peristalsis \cite{meng2004wireless}. 
Therefore, \textit{Active WCE}, which is a new concept of endowing a robotic capsule with active locomotion and precise localization, holds great promise to overcome the drawbacks
of conventional endoscopy to shorten the examination time, relieve operator burden and improve the accuracy of diagnosis and therapy \cite{meng2004wireless}\cite{ciuti2011capsule}.

In recent years, simultaneous magnetic actuation and localization (SMAL) for active WCE have become an active area of research, and a large number of systems have been developed based on electromagnets or permanent magnets to propel and locate the capsule in the intestine \cite{bianchi2019localization}\cite{shamsudhin2017magnetically}\cite{abbott2020magnetic}\cite{kummer2010octomag}. Compared with the electromagnet-based systems, the permanent magnet-based systems generally have the advantages of being more compact, affordable and energy-efficient, and having a larger workspace \cite{pittiglio2019magnetic}.
In the permanent magnet-based systems, a capsule with embedded magnet(s) is actuated by an external permanent magnet, and their magnetic fields are measured by magnetic sensors to locate the capsule.

Some permanent magnet-based systems directly use the magnetic force to drag the capsule to track a manually specified trajectory \cite{mahoney2016five}\cite{taddese2018enhanced}\cite{barducci2019adaptive}\cite{scaglioni2019explicit} or automatically explore an unknown environment \cite{martin2020enabling}.
Instead of exclusively utilizing the magnetic force for dragging, Mahoney et al. \cite{mahoney2014generating} and Popek et al. \cite{popek2017first}\cite{popek2016six} used a continuously rotating spherical-magnet actuator \cite{wright2015spherical} to generate a rotating magnetic field for helical propulsion of a capsule in a tubular environment, which we refer to as \textit{continuously rotating magnetic actuation} (CRMA).
Other researchers also  employed CRMA for capsule actuation in several external sensor array-based SMAL systems to propel a capsule in the intestinal environment \cite{xu2020novelsystem}\cite{xu2019towards}\cite{xu2020novel}\cite{xu2020improved}.
However, since the small intestine is a soft and narrow tubular environment with an inner diameter of only about $2 cm$ \cite{gray1974anatomy}, and the intestinal peristalsis involves periodic contraction of intestinal muscles \cite{elaine2017essentials}, it is of concern that the continuous rotation of the capsule with an external screw thread may result in harmful deformation of the intestine or even volvulus due to the rotational motion of the capsule, which has been reported in prior arts \cite{xu2020novelsystem}.
Therefore, although existing CRMA methods have realized efficient propulsion of the capsule, the risk of causing harmful deformation of the intestine such as volvulus due to the rotation of the capsule needs to be considered and analyzed to improve patient safety.

\begin{figure*}[t]
\setlength{\abovecaptionskip}{-0.0cm}
\centering
\includegraphics[scale=1.0,angle=0,width=15cm]{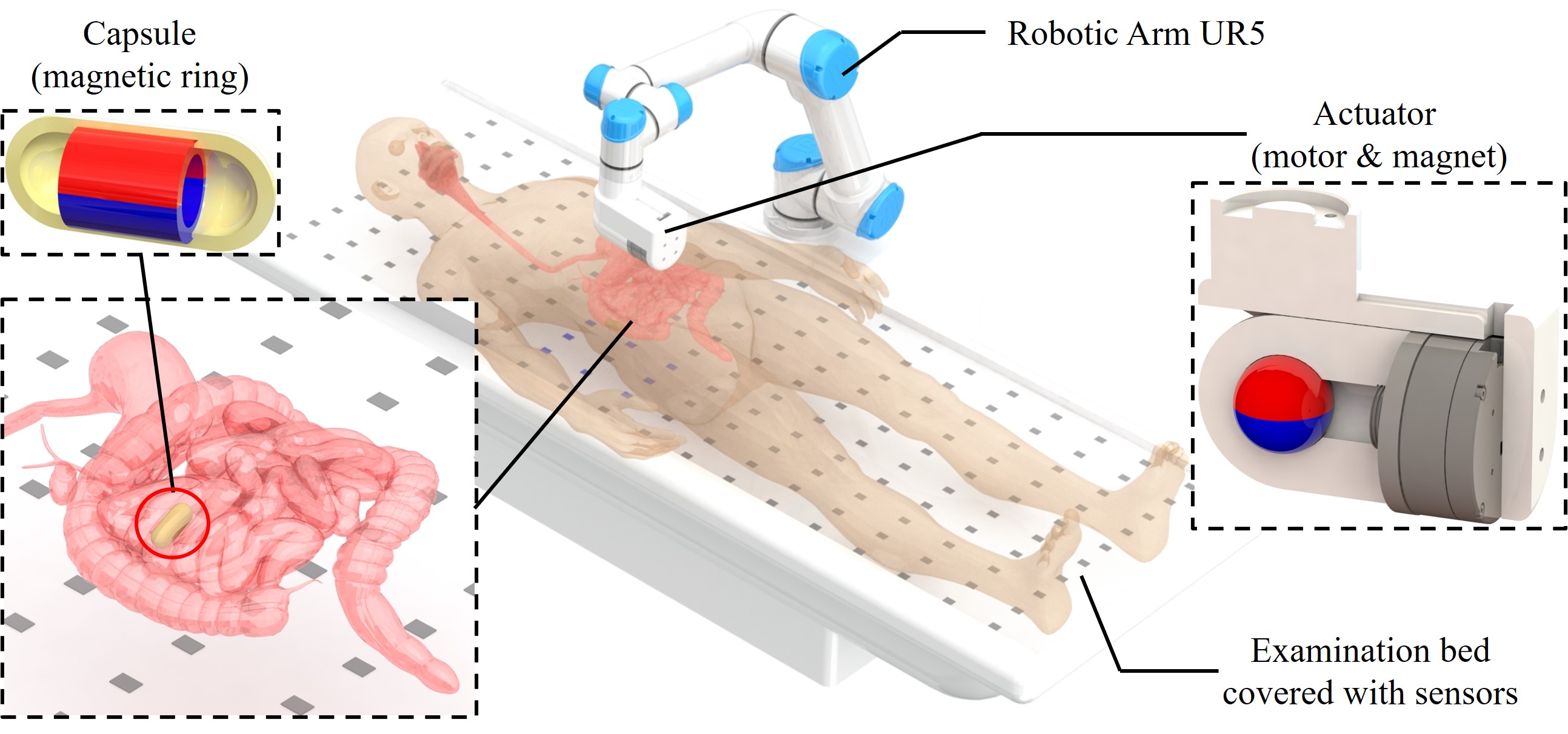}
\caption{The overall design and application scenario of our SMAL system, which includes a reciprocally rotating actuator magnet mounted at the end-effector of a robotic arm and a robotic capsule embedded with a magnetic ring. The patient is required to swallow the capsule and lie in the supine position on an examination bed covered with magnetic sensors. The actuator magnet rotates reciprocally above the capsule to automatically propel it through the intestine.}
\label{Fig_system3D}
\end{figure*}

In this work, in order to mitigate the risk of causing damage to the intestinal wall and enhance patient safety in the WCE application, we forgo the external screw thread of the capsule and use the magnetic force to actuate the capsule. Moreover, we employ a rotating magnetic field to actuate the capsule based on the assumption that the rotational movement can help make the intestine stretch open for easier advancement of the capsule. We here propose a novel approach for rotating magnetic actuation, namely, the \textit{reciprocally rotating magnetic actuation} (RRMA), in order to reduce the effect of causing volvulus or malrotation of the intestine and improve patient safety. The actuation model for RRMA is formulated to calculate the desired actuator pose to generate a desired rotating magnetic field, and a theoretical analysis of the potential risk of causing volvulus due to the capsule motion is provided.
Moreover, we integrate the RRMA method with existing magnetic localization algorithms to realize closed-loop SMAL of the capsule.
The overall SMAL system with its application scenario is illutrated in Fig. \ref{Fig_system3D}, which is developed based on our prototypes presented in \cite{xu2020novelsystem}\cite{xu2020improved}.

The main contributions of this paper are three-fold:
\begin{itemize}
\item A novel actuation approach for active WCE named RRMA is proposed as an alternative to the conventional CRMA to enhance patient safety. We formulate the actuation model and show how to control the actuator to generate a desired rotating magnetic field.
\item The risk of causing malrotation of the intestine or volvulus under CRMA and RRMA is theoretically analyzed and qualitatively demonstrated to show the potential benefit of using RRMA for capsule actuation.
\item We implement the proposed RRMA approach on a real robotic system and demonstrate the effectiveness of the method in the automatic propulsion of a capsule in an ex-vivo pig colon.
\end{itemize}

Although the work presented in this paper is targeted at the automatic propulsion of a robotic capsule in the intestinal environment, the methods and ideas described herein can be applied to the magnetic actuation of robots working in tubular environments in general.

The remainder of this paper is organized as follows. Section II presents the nomenclature of this paper. The details of our proposed RRMA method is introduced in Section III. Experiment results are presented in Section IV, before we discuss and conclude this work in Section V.

\section{Nomenclature}

Throughout this paper, lowercase normal fonts refer to scalars (e.g., $\mu_{0}$). Lowercase bold fonts refer to vectors (e.g., $\textbf{b}$). The vector with a ``hat'' symbol indicates that the vector is a unit vector of the original vector (e.g., $\widehat{\textbf{r}}$ is the unit vector of $\textbf{r}$), and the vector with an ``[index]'' indicates one component of the vector (e.g., $\textbf{r}[i]$ is the $i$-th component of $\textbf{r}$, $\textbf{r}\in\mathbb{R}^{n\times1}$, $1 \leq i \leq n$). Matrices are represented by uppercase bold fonts (e.g. $\textbf{M}$), and $\textbf{I}_{n}$ denotes $n \times n$ identity matrix. In addition, $Rot_{k}(\theta)$ represents the rotation of $\theta$ degrees around the $+k$-axis, $k\in\{x,y,z\}$.

\section{Reciprocally Rotating Magnetic Actuation}

In this section, we first introduce the RRMA model to show how to control the pose of the actuator to generate a desired reciprocally rotating magnetic field for capsule actuation, and then provide a theoretical analysis of the potential risk of causing volvulus due to the capsule motion under CRMA and RRMA, respectively. Finally, the RRMA-based SMAL workflow is presented to achieve automatic propulsion of a robotic capsule in an unknown tubular environment.

\subsection{Reciprocally Rotating Magnetic Actuation Model}

\begin{figure*}[t]
\setlength{\abovecaptionskip}{-0.2cm}
\centering
\includegraphics[scale=1.0,angle=0,width=1.0\textwidth]{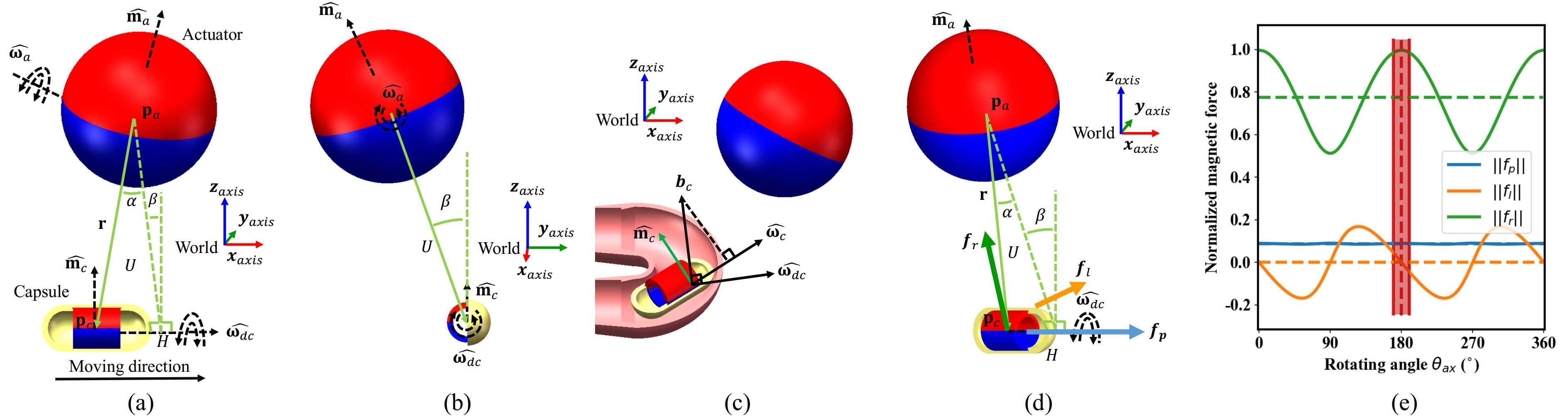}
\caption{(a) The capsule which reciprocally rotates around $\widehat{\pmb{\omega}_{dc}}$ at $\mathbf{p}_{c}$ is actuated by an actuator rotating around $\widehat{\pmb{\omega}_{a}}$ at $\mathbf{p}_{a}$. $\mathbf{p}_{a} H \perp \widehat{\pmb{\omega}_{dc}}$. $\mathbf{r}$ is the vector from $\mathbf{p}_{a}$ to $\mathbf{p}_{c}$. $U$ is the plane formed by $\mathbf{p}_{a}$, $\mathbf{p}_{c}$ and $H$. $\alpha$ is the angle between $\mathbf{r}$ and the $\mathbf{p}_{a} H$, and $\beta$ is the angle between plane $U$ and the vertical line. (b) is the side view of (a). (c) The direction of $\widehat{\mathbf{m}_{c}}$ is jointly determined by the magnetic field $\mathbf{b}_{c}$ and the current rotation axis of the capsule $\widehat{\pmb{\omega}_{c}}$. (d) The propulsive force $\mathbf{f}_{p}$ is along the direction of $\widehat{\pmb{\omega}_{dc}}$, and the lateral force $\mathbf{f}_{l}$ is perpendicular with plane $U$. (e) shows the normalized propulsive force, lateral force and the remainder force applied to the capsule over one actuator revolution ($\alpha=10^{\circ}$, $\beta=0^{\circ}$), and the dashed lines refer to the corresponding average values (the blue dashed line is obscured by the blue solid line). The reciprocal rotation range is illustrated as the red shadowed region, and the red dashed line is the center of the reciprocal rotation range.}
\label{Fig_RRMA_model}
\end{figure*}

As shown in Fig. \ref{Fig_RRMA_model}, a robotic capsule embedded with a passive magnet ring is actuated in a magnetic field generated by a reciprocally rotating actuator magnet with a spherical shape. Assume that the desired rotation axis of the capsule $\widehat{\pmb{\omega}_{dc}}$ is initially aligned with $+x$-axis of the world frame, then $\widehat{\pmb{\omega}_{dc}}$ can be represented by $\theta_{cz}$, $\theta_{cy}$ using (\ref{F_wc}).

\begin{equation}
\label{F_wc}
\begin{aligned}
&\quad\ \widehat{\pmb{\omega}_{dc}} = Rot_{z}(\theta_{cz})Rot_{y}(-\theta_{cy})\left(\begin{matrix}1\\0\\0\end{matrix}\right)\\
\Longrightarrow &
\begin{cases}
\begin{split}
\theta_{cz} &= \arctan{\frac{\widehat{\pmb{\omega}_{dc}}[2]}{\widehat{\pmb{\omega}_{dc}}[1]}}, &\theta_{cz} \in [0^{\circ}, 360^{\circ})\quad \\
\theta_{cy} &= \arcsin{\widehat{\pmb{\omega}_{dc}}[3]}, &\theta_{cy} \in (-90^{\circ}, 90^{\circ})
\end{split}
\end{cases}
\end{aligned}
\end{equation}

Assume that the unit magnetic moment of the actuator $\widehat{\mathbf{m}_{a}}$ is initially aligned with $+z$-axis of the world frame, let $\theta_{ax} \in [0^{\circ}, 360^{\circ})$ indicate the angle that $\widehat{\mathbf{m}_{a}}$ rotates around $\widehat{\pmb{\omega}_{a}}$, then $\widehat{\mathbf{m}_{a}}$ can be calculated by (\ref{F_ma}).

\begin{equation}
\label{F_ma}
\widehat{\mathbf{m}_{a}}=Rot_{z}(\theta_{az})Rot_{y}(-\theta_{ay})Rot_{x}(\theta_{ax})\left(\begin{matrix}0\\0\\1\end{matrix}\right)
\end{equation}

In order to generate a rotating magnetic field around the desired moving direction of the capsule $\widehat{\pmb{\omega}_{dc}}$ at $\mathbf{p}_{c}$, the rotation axis of the actuator $\widehat{\pmb{\omega}_{a}}$ can be calculated by (\ref{F_rotating_actuation}) \cite{mahoney2014generating}.

\begin{equation}
\label{F_rotating_actuation}
\widehat{\pmb{\omega}_{a}}=\Widehat{\left((3\widehat{\mathbf{r}}{\widehat{\mathbf{r}}}^{T}-\mathbf{I}_{3})\ \widehat{\pmb{\omega}_{dc}}\right)}
\end{equation}

\noindent where $\mathbf{r}=\mathbf{p}_{c}-\mathbf{p}_{a}$ is the position of the capsule relative to the actuator. As illustrated in Fig. \ref{Fig_RRMA_model}(a)(b), let $H$ be the projection of $\mathbf{p}_{a}$ onto the direction of $\widehat{\pmb{\omega}_{dc}}$, let $\alpha$ represent the angle between $\mathbf{r}$ and $\mathbf{p}_{a} H$, and $\beta$ represents the angle between plane $U$ formed by $\mathbf{p}_{a}$, $\mathbf{p}_{c}$ and $H$ and the vertical line. Assume the distance between the capsule and the actuator is $d$, then $\mathbf{r}=d\cdot \widehat{\mathbf{r}}$ can be represented by (\ref{F_r}).


\begin{equation}
\label{F_r}
\mathbf{r}=d\left(Rot_{z}(\theta_{cz})Rot_{y}(-\theta_{cy})Rot_{x}(\beta)Rot_{y}(\alpha)\left(\begin{matrix}0\\0\\-1\end{matrix}\right)\right)
\end{equation}


Assume that the rotation axis of the actuator $\widehat{\pmb{\omega}_{a}}\in\mathbb{R}^{3 \times 1}$ is initially aligned with $+x$-axis of the world frame, then $\widehat{\pmb{\omega}_{a}}$ can be represented by $\theta_{az}$, $\theta_{ay}$ using (\ref{F_wa}).

\begin{equation}
\label{F_wa}
\begin{aligned}
&\quad\ \widehat{\pmb{\omega}_{a}} = Rot_{z}(\theta_{az})Rot_{y}(-\theta_{ay})\left(\begin{matrix}1\\0\\0\end{matrix}\right)\\
\Longrightarrow &
\begin{cases}
\begin{split}
\theta_{az} &= \arctan{\frac{\widehat{\pmb{\omega}_{a}}[2]}{\widehat{\pmb{\omega}_{a}}[1]}}, &\theta_{az} \in [0^{\circ}, 360^{\circ})\quad \\
\theta_{ay} &= \arcsin{\widehat{\pmb{\omega}_{a}}[3]}, &\theta_{ay} \in (-90^{\circ}, 90^{\circ})
\end{split}
\end{cases}
\end{aligned}
\end{equation}

Let $\|\mathbf{m}_{a}\|$ denote the magnitude of the magnetic moment of the actuator, which is influenced by the material and the geometry of the actuator magnet, then the rotating magnetic field applied to the capsule $\mathbf{b}_{c}$ can be calculated according to the magnetic dipole model \cite{cheng1989field}:


\begin{equation}
\label{F_bc}
\mathbf{b}_{c}(\mathbf{r},\widehat{\mathbf{m}_{a}})=\frac{\mu_{0}\|\mathbf{m}_{a}\|}{4\pi\|\mathbf{r}\|^5}\left(3\mathbf{r}\mathbf{r}^{T}-(\mathbf{r}^{T}\mathbf{r})\mathbf{I}_{3}\right)\widehat{\mathbf{m}_{a}}
\end{equation}


\begin{figure*}[t]
\setlength{\abovecaptionskip}{-0.1cm}
\centering
\includegraphics[scale=1.0,angle=0,width=1.0\textwidth]{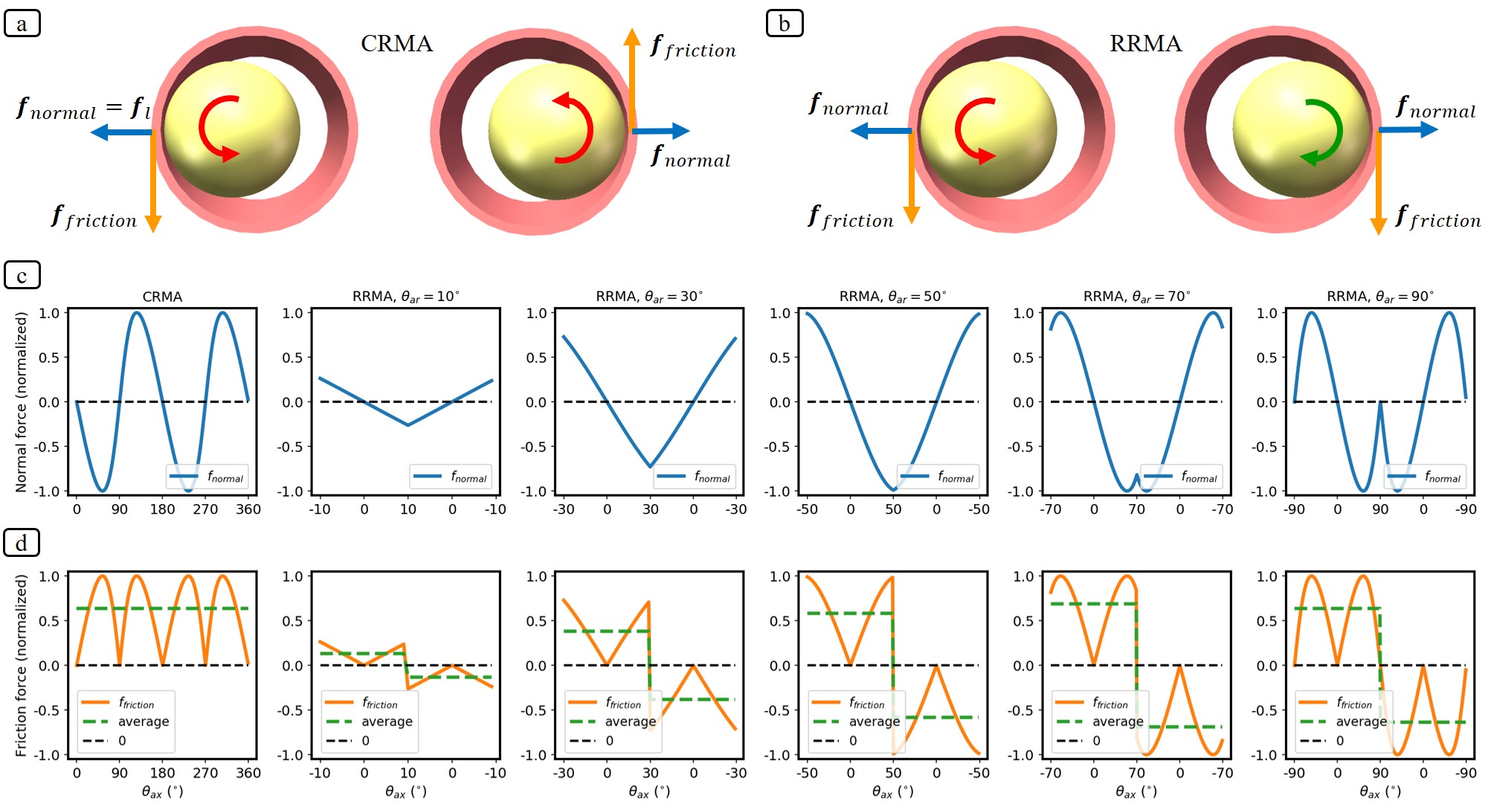}
\caption{(a) and (b) illustrate the normal force $\mathbf{f}_{normal}$ and friction $\mathbf{f}_{friction}$ applied to the intestine by the capsule under CRMA and RRMA, respectively. The normal force $\mathbf{f}_{normal}$ is assumed to be equal to the lateral magnetic force $\mathbf{f}_{l}$ experienced by the capsule. Under CRMA, $\mathbf{f}_{friction}$ always causes the intestine to twist along one direction, while under RRMA, the twisting direction of the intestine will periodically change to the opposite direction. (c) and (d) plot the change of $\mathbf{f}_{normal}$ and $\mathbf{f}_{friction}$ with respect to $\theta_{ax}$ in one cycle of the actuator's movement under CRMA and RRMA with different reciprocating rotation angles $\theta_{ar}=10^{\circ}$, $30^{\circ}$, $50^{\circ}$, $70^{\circ}$, $90^{\circ}$.}
\label{Fig_force_distribution}
\end{figure*}

We do not assume that the actual magnetic moment of the capsule is always aligned with the magnetic field (i.e., $\widehat{\mathbf{m}_{c}}=\widehat{\mathbf{b}_{c}}$, as would be the case in liquid filled cavities \cite{mahoney2016five}), because the movement of the capsule is constrained by the inner wall of the narrow tubular environment, as shown in Fig. \ref{Fig_RRMA_model}(c). Instead, we assume that the direction of $\widehat{\mathbf{m}_{c}}$ is jointly determined by the magnetic field $\mathbf{b}_{c}$ and the current rotation axis of the capsule $\widehat{\pmb{\omega}_{c}}$, which can be estimated from the measurement of magnetic sensors \cite{xu2020improved}: 


\begin{equation}
\label{F_mc}
\widehat{\mathbf{m}_{c}}=\Widehat{(\mathbf{b}_{c}-(\mathbf{b}_{c}^{T}\widehat{\pmb{\omega}_{c}})\widehat{\pmb{\omega}_{c}})}
\end{equation}


The magnetic force $\mathbf{f}$ applied to the capsule can be calculated using (\ref{F_f}):


\begin{equation}
\label{F_f}
\begin{aligned}
\mathbf{f}&=\mathbf{f}(d,\alpha,\beta,\theta_{ax},\widehat{\pmb{\omega}_{dc}})=\mathbf{f}(\mathbf{r},\widehat{\mathbf{m}_{a}},\widehat{\mathbf{m}_{c}},\widehat{\pmb{\omega}_{dc}})\\
&=\frac{3\mu_{0}||\mathbf{m}_{a}||\cdot||\mathbf{m}_{c}||}{4\pi||\mathbf{r}||^{7}}\left((\widehat{\mathbf{m}_{c}}\widehat{\mathbf{m}_{a}}^{T})\mathbf{r}(\mathbf{r}^{T}\mathbf{r})\right.\\
&\left.+(\widehat{\mathbf{m}_{a}}\widehat{\mathbf{m}_{c}}^{T})\mathbf{r}(\mathbf{r}^{T}\mathbf{r})+(\widehat{\mathbf{m}_{c}}^{T}(\mathbf{I}_{3}\mathbf{r}^{T}\mathbf{r}-5\mathbf{r}\mathbf{r}^{T})\widehat{\mathbf{m}_{a}})\mathbf{r}\right)\\
\end{aligned}
\end{equation}


As shown in Fig. \ref{Fig_RRMA_model}(d), $\mathbf{f}$ can be resolved into three components: (i) the propulsive force $\mathbf{f}_{p}$, which is the projection of $\mathbf{f}$ on the desired heading direction of the capsule $\widehat{\pmb{\omega}_{dc}}$, (ii) the lateral force $\mathbf{f}_{l}$, which is perpendicular with plane $U$, and (iii) the remainder force $\mathbf{f}_{r}=\mathbf{f}-\mathbf{f}_{p}-\mathbf{f}_{l}$. In order to determine the center of the reciprocating rotation of the actuator for RRMA, we plot the magnetic force applied to the capsule over one actuator revolution in Fig. \ref{Fig_RRMA_model}(e). $\|\mathbf{f}_{p}\|$ changes little with $\theta_{ax}$. When $\theta_{ax}$ is around $0^{\circ}$ or $180^{\circ}$, $\|\mathbf{f}_{l}\|$ is around zero, and $\|\mathbf{f}_{r}\|$ reaches its peak. This indicates that $\mathbf{f}$ changes slowly around $\theta_{ax}=180^{\circ}$. Let $\theta_{ar}$ denotes the angle of the actuator's reciprocal rotation, i.e., $\theta_{ax} \in [180^{\circ}-\theta_{ar},180^{\circ}+\theta_{ar}]$. When $\theta_{ar}$ is relatively small, $\mathbf{f}$ in (\ref{F_f}) can be approximated using (\ref{F_f_theta_c_approximate}) in the practical application.

\begin{equation}
\label{F_f_theta_c_approximate}
\begin{split}
\mathbf{f}(d,\alpha,\beta,\theta_{ax},\ &\widehat{\pmb{\omega}_{dc}}) \doteq \mathbf{f}(d,\alpha,\beta,\theta_{ax}=180^{\circ},\widehat{\pmb{\omega}_{dc}}),\\
\theta_{ax} &\in [180^{\circ}-\theta_{ar},180^{\circ}+\theta_{ar}]
\end{split}
\end{equation}

\subsection{Risk of Causing Volvulus with Different Actuation Modes} \label{section_volvulus}

In order to illustrate the potential benefit of RRMA over CRMA and determine the best reciprocating angle, we qualitatively analyze the risk of causing intestinal malrotation and volvulus with different actuation modes by taking a look at the forces exerted onto the intestinal wall by the capsule in some simplified cases, as shown in Fig. \ref{Fig_force_distribution}.
Without loss of generality, we set $\alpha=0^{\circ}$, $\beta=0^{\circ}$, and assume that the capsule rotates in a fixed tubular intestine with an inner diameter slightly larger than the capsule's diameter.

As shown in Fig. \ref{Fig_force_distribution}(a), when the capsule is pressed against the left side of the intestinal wall by the lateral magnetic force $\mathbf{f}_{l}$, the intestinal wall will experience a normal force $\mathbf{f}_{normal} = \mathbf{f}_{l}$ and a friction force $\mathbf{f}_{friction}$ ($\|\mathbf{f}_{friction}\| \propto \|\mathbf{f}_{normal}\|$), which may cause the intestine to twist in the counterclockwise direction. When the capsule touches the right side of the intestinal wall under CRMA, the upward friction force exerted upon the intestinal wall by the capsule may also twist the intestine in the counterclockwise direction. Therefore, if the capsule continuously rotates in the narrow intestine, the net twisting may bring a higher risk of causing intestinal malrotation and volvulus. 
In contrast, as shown in Fig. \ref{Fig_force_distribution}(b), when the capsule is pressed against the left intestinal wall, the friction may also twist the intestine in the counterclockwise direction. But when the capsule rotates in the opposite direction and touches the right side under RRMA, the downward friction force will untwist the intestine, thereby mitigating the risk of volvulus.
Fig. \ref{Fig_force_distribution}(c) illustrates the normal force (normalized) under CRMA and RRMA with different reciprocating rotation angles (i.e, $\theta_{ar}=10^{\circ}$, $30^{\circ}$, $50^{\circ}$, $70^{\circ}$, $90^{\circ}$) in one cycle of the actuator's movement. Fig. \ref{Fig_force_distribution}(d) shows the friction force (normalized), where a positive value means the friction in the counterclockwise direction. As shown in the first column, CRMA always has the tendency to twist the intestine in the same direction. However, as shown in the last five columns, although the average magnitude of friction over a half reciprocation cycle increases as $\theta_{ar}$ gets larger, the opposite effects of twisting the intestine cancel each other out in one cycle. Therefore, the reciprocating rotation mode proposed in RRMA can potentially prevent intestinal malrotation and enhance patient safety during the examination. 

In addition, when the capsule is moving in the narrow intestine, the higher hoop stress at the front of the capsule where the lumen is stretching open will result in a greater environmental resistance \cite{zhang2012modeling}\cite{incetan2020vr}. As shown in Fig. \ref{Fig_forcesensor}(a)(b), the average values of $\|f_{normal}\|$ and $\|f_{friction}\|$ over a half cycle of reciprocation gradually increase as the angle of reciprocating rotation $\theta_{ar}$ changes from $10^{\circ}$ to $90^{\circ}$. A larger normal force applied to the intestinal wall by the capsule $f_{normal}$ during rotation will have a positive effect to make the lumen stretch open and overcome the environmental resistance. Therefore, we choose $90^{\circ}$ as the reciprocating rotation angle used in our RRMA approach.

\subsection{Workflow of the RRMA-based SMAL System}

\begin{figure}[t]
\centering
\includegraphics[scale=1.0,angle=0,width=0.31\textwidth]{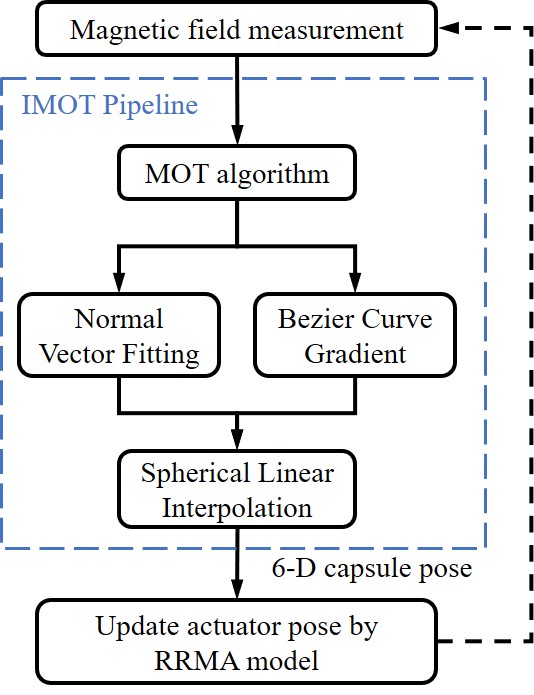}
\caption{Workflow of the RRMA-based SMAL system, which uses the improved multiple objects tracking (IMOT) pipeline in \cite{xu2020improved} (in the blue dashed rectangle) to estimate the 6-DOF pose of the capsule from magnetic field measurements and updates the actuator pose by the proposed RRMA model.}
\label{Fig_RRMA_SMAL_workflow}
\end{figure}

In order to implement the RRMA method on a real robotic system to realize closed-loop SMAL of the capsule and apply the system in the automatic propulsion task, we integrate the RRMA-based actuation method with our previously proposed \textit{improved multiple objects tracking} (IMOT) algorithm \cite{xu2020improved} for capsule localization to build a closed-loop SMAL system. The workflow of the system is shown in Fig. \ref{Fig_RRMA_SMAL_workflow}.

Given the magnetic field measurements provided by the magnetic sensors, the IMOT pipeline first uses the multiple objects tracking (MOT) algorithm \cite{song2016multiple} to estimate the 5-D pose of the capsule, then two 6-D pose estimations are separately calculated using a normal vector fitting-based method and a B\'{e}zier curve gradient-based method, before the final 6-D pose estimation ($\mathbf{p}_{c}, \widehat{\mathbf{m}_{c}}, \widehat{\pmb{\omega}_{c}}$) is generated by combining the results with spherical linear interpolation. Details of this localization algorithm is described in \cite{xu2020improved}. 

Similar to \cite{xu2020novelsystem}\cite{xu2019towards}\cite{xu2020novel}, in this work, we consider the task of automatically propelling the capsule in an unknown tubular environment. Therefore, we set the desired rotation axis of the capsule $\widehat{\pmb{\omega}_{dc}}$ in (\ref{F_rotating_actuation}) to be along the current moving direction of the capsule $\widehat{\pmb{\omega}_{c}}$, which is estimated in real time by the IMOT-based localization algorithm, and the position of the actuator relative to the capsule is kept unchanged by setting $\alpha$, $\beta$ and $d$ in (\ref{F_r}) to fixed values. Then, the actuator pose will be updated based on the RRMA method introduced in Section~III-A.

\begin{figure}[t]
\centering
\includegraphics[scale=1.0,angle=0,width=0.43\textwidth]{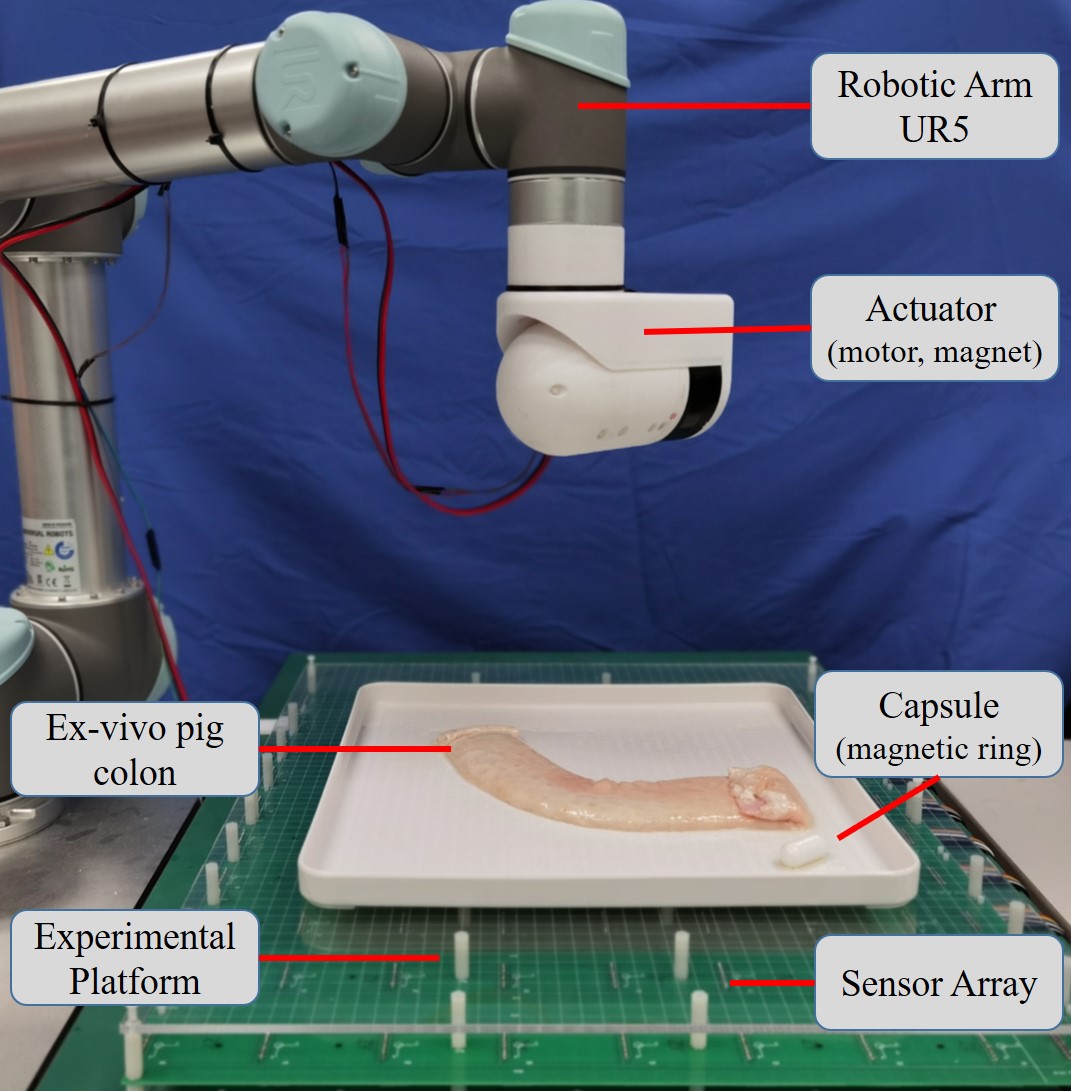}
\caption{System setup for the validation of our RRMA method during the experiments in an ex-vivo pig colon.}
\label{Fig_implementation_system}
\end{figure}

\section{Experiments and Results}

\subsection{System Setup}

In order to evaluate the performance of the proposed method, we build a SMAL system based on the prototypes presented in our previous works \cite{xu2020novelsystem}\cite{xu2020improved} and according to the system design in Fig. \ref{Fig_system3D}. The experimental setup is illustrated in Fig. \ref{Fig_implementation_system}. A 6-DoF serial manipulator (5-kg payload, UR5, Universal Robots) is used to control the movement of the actuator, which consists of a DC motor (RMD-L-90, GYEMS) and a spherical permanent magnet (diameter $50mm$, NdFeB, N42 grade). The capsule (diameter $16mm$, length $35mm$) consists of a 3D-printed shell (Polylactic Acid, UP300 3D printer, Tiertime) and a permanent magnetic ring (outer diameter $12.8mm$, inner diameter $9mm$, length $15mm$, NdFeB, N38SH grade) inside. The external sensor array includes $80$ three-axis magnetic sensors (MPU9250, InvenSense) arranged in an $8\times10$ grid with a spacing of $6cm$. The output frequency of each sensor is $100Hz$. $10$ USB-I2C adaptors (Ginkgo USB-I2C, Viewtool), a USB-CAN adaptor (Ginkgo USB-CAN, Viewtool) and a network cable are used for data transmission. 

The proposed RRMA approach is implemented with Python on a desktop (Intel i7-7820X, 32GB RAM, Win10).  The real-time localization of the capsule used in this system can achieve a localization accuracy of $4.3\pm1.9 mm$ and $5.4\pm1.7^{\circ}$ in position and orientation, respectively.

\subsection{{Evaluation of Lateral Force under RRMA}}

\begin{figure}[t]
\setlength{\abovecaptionskip}{-0.2cm}
\centering
\includegraphics[scale=1.0,angle=0,width=0.49\textwidth]{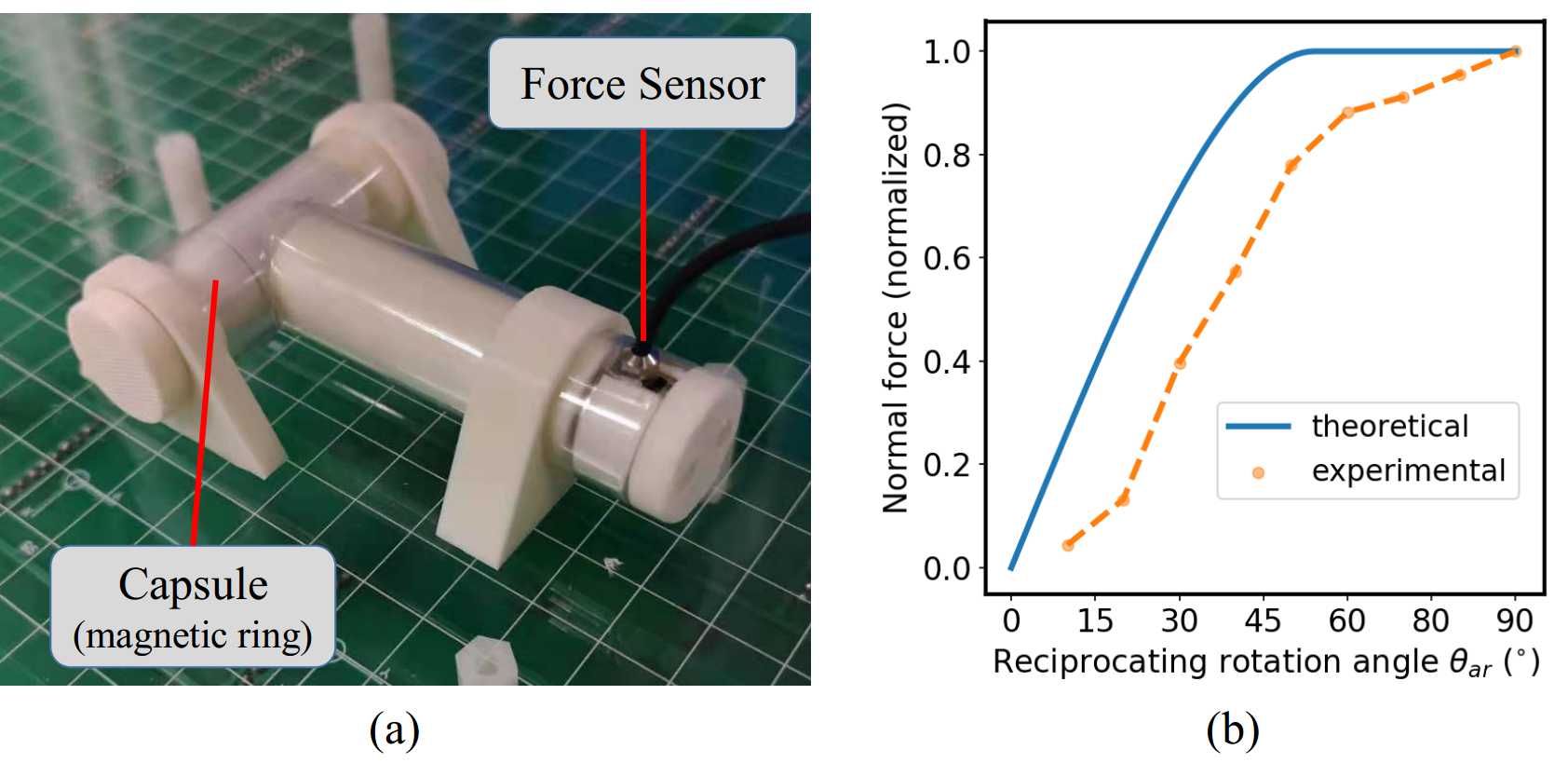}
\caption{(a) The lateral magnetic forces experience by the capsule with different reciprocating rotation angles are measured in a ``T"-shaped tube. (b) shows the maximum magnitudes of the theoretical normal forces and the experimental results.}
\label{Fig_forcesensor}
\end{figure}

First, we conduct an experiment to measure the lateral force experienced by a reciprocally rotating capsule when different reciprocating rotation angles $\theta_{ar}$ are applied in the RRMA method. As shown in Fig. \ref{Fig_forcesensor}(a), the capsule is actuated in a ``T"-shaped PVC tube, and a force sensor (AT8600A-5N, AUTODA) is rigidly attached to a PLA cylinder which is in contact with the capsule to measure the normal force applied by the capsule on the tube under RRMA. 

As shown in Fig. \ref{Fig_forcesensor}(b), when the reciprocating rotation angle $\theta_{ar}$ is chosen between $0 ^{\circ}$ and $90 ^{\circ}$, the maximum magnitudes of the theoretical normal forces (as analyzed in Section~\ref{section_volvulus}) and the measurement results are plotted with blue and orange curves, respectively. It can be observed that as the reciprocating rotation angle $\theta_{ar}$ becomes larger, the change of the measured normal force is basically consistent with the theoretical value, but slightly lags behind the theoretical value. This may be because the magnetic moment of the capsule is not changed by the applied magnetic field in real time due to the friction of the environment.

\subsection{{Evaluation of the Risk of Causing Volvulus under CRMA and RRMA}}

\begin{figure}[t]
\centering
\includegraphics[scale=1.0,angle=0,width=0.48\textwidth]{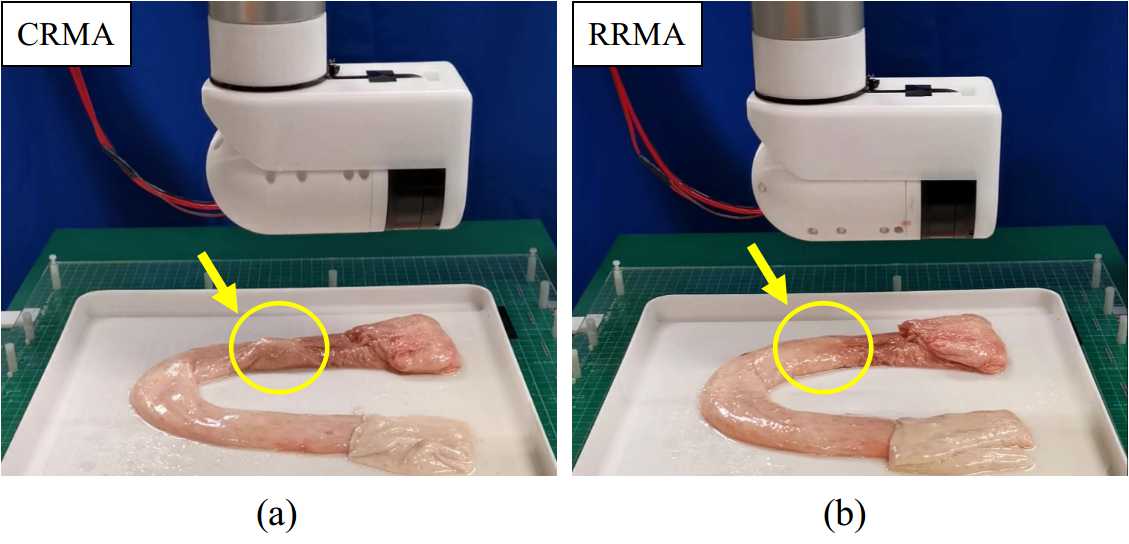}
\caption{The capsule is actuated in an ex-vivo pig colon under different rotating magnetic actuation modes. In (a), the continuously rotating capsule under CRMA can occasionally cause the intestine to twist, which may bring a high risk of causing intestinal malrotation and volvulus, while in (b), the reciprocally rotating capsule under RRMA can maintain the normal shape of the intestine, which has the potential to improve patient safety.}
\label{Fig_compare_CRMA_RRMA}
\end{figure}

Then, we qualitatively evaluate the risk of causing volvulus by using the conventional CRMA \cite{mahoney2014generating} and the proposed RRMA in two capsule actuation experiments conducted in an ex-vivo pig colon, as shown in Fig. \ref{Fig_compare_CRMA_RRMA}. In both experiments, the actuator is located $13cm$ above the capsule and rotates at a speed of $360^{\circ}/sec$. 

As shown in Fig. \ref{Fig_compare_CRMA_RRMA}(a), when CRMA is applied to actuate the capsule, it can be seen that the continuously rotating motion of the capsule occasionally causes the intestine to twist, which may result in the intestinal malrotation and even volvulus. This situation should be avoided, as it may cause patient discomfort and even result in severe patient injury in clinical applications. In contrast, no twist or abnormal deformation of the intestine is observed during the actuation experiment using our proposed RRMA method, as shown in Fig. \ref{Fig_compare_CRMA_RRMA}(b). These results qualitatively demonstrate the effectiveness of our method to mitigate the risk of causing intestinal malrotation and volvulus during the capsule actuation compared with the CRMA method. Therefore, the proposed RRMA approach has the potential to improve the clinical acceptability of the technology by reducing risk and enhancing patient safety.

\subsection{{Evaluation of Different Actuation Methods in the Automatic Propulsion Task}}

To further demonstrate the effectiveness of the proposed RRMA method, we apply the SMAL system to the automatic propulsion task (as described in Section~III-C) using different actuation methods, i.e., dragging magnetic actuation (DMA), continuously rotating magnetic actuation (CRMA), and reciprocally rotating magnetic actuation (RRMA). The capsule is automatically propelled to move through an ex-vivo pig colon with a length of $155mm$. We carried out five instances for each test. The actuating angle $\alpha$ is set to $10^{\circ}$, the distance between the actuator and the capsule $d$ is set to $15 cm$, and $\beta$ is set to zero for all the experiments. These values are set empirically through a number of experiments to achieve a balance between the actuation speed and localization accuracy, as well as to keep a safe distance between the actuator and the patient.

The experiment results are presented in Table \ref{T_AP}. It is found that the capsule cannot be effectively propelled in the pig colon by DMA due to the large friction in the environment. In contrast, the rotating actuation based methods (i.e., CRMA and RRMA) can reduce the friction between the capsule and the colon wall to accelerate the exploration in the unknown tubular environment with an average moving speed of over $2mm/s$. While the CRMA method can successfully propel the capsule through the pig colon in $40\%$ of the five experiments, the RRMA method reaches the highest success rate of $100\%$ with an average speed of $2.48 mm/s$. As observed in our experiments, the capsule actuated by CRMA occasionally causes malrotation of the intestinal wall, which would increase the environmental resistance and hinder the movement of the capsule, and more importantly, cause damage to the intestinal tissues. In contrast, the RRMA method does not cause the twist of the intestine during all the experiments and can smoothly propel the capsule though the pig colon, which shows its potential to improve patient safety and actuation efficiency. These results show that the RRMA-based SMAL system can realize efficient and robust propulsion of the capsule in an unknown tubular environment.

\begin{table}[tb] \renewcommand\arraystretch{1.2} \small
\centering
\caption{{Performance of the Three Magnetic Actuation Methods in the Automatic Propulsion Task}}
\begin{tabular}{|c|c|c|}
\hline
 \multirow{1}{*}{\tabincell{c}{Method}} & \multirow{1}{*}{Success rate} & \multirow{1}{*}{\tabincell{c}{Average moving speed}} \\
\hline
 DMA & {$0 \%$} & {$-$} \\
\hline {CRMA} & {$40 \%$} & {$2.58$  $mm/s$} \\
\hline {RRMA} & {$100 \%$} & {$2.48$ $mm/s$} \\
\hline
\end{tabular}
\label{T_AP}
\end{table}

\begin{figure*}[t]
\centering
\includegraphics[scale=1.0,angle=0,width=0.85\textwidth]{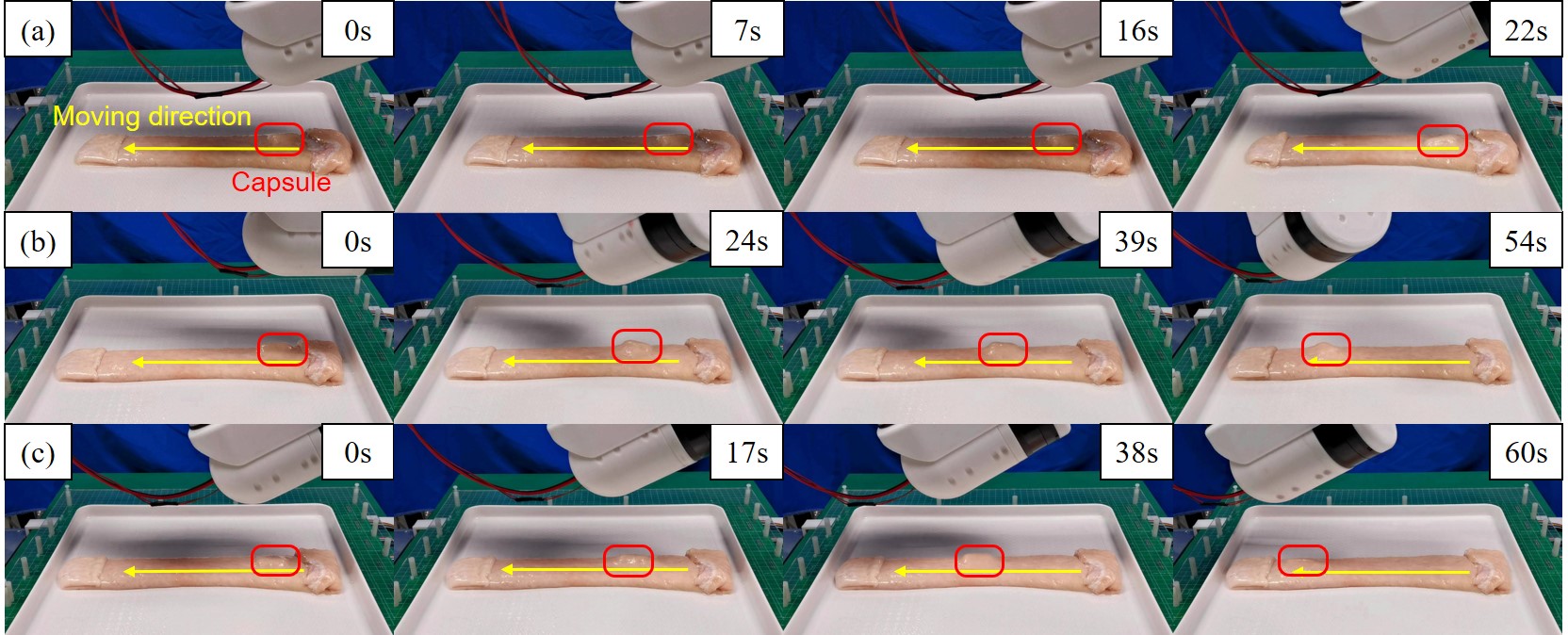}
\caption{{Snapshots of the trajectories of the capsule in automatic propulsion experiments in an ex-vivo pig colon using different magnetic actuation methods (a) dragging magnetic actuation (DMA), (b) continuously rotating magnetic actuation (CRMA), and (c) reciprocally rotating magnetic actuation (RRMA). The moving direction of the capsule is illustrated by yellow arrows, and the position of the capsule is highlighted in red rectangles.}}
\label{Fig_experiment_3MA}
\end{figure*}

We further compare the performance of the three magnetic actuation methods through qualitative analysis. The moving trajectories of the capsule under three actuation modes in example cases are illustrated in Fig. \ref{Fig_experiment_3MA}. The DMA method cannot overcome the large friction to effectively propel the capsule (see Fig. \ref{Fig_experiment_3MA}(a)), while the rotating actuation based methods can successfully propel the capsule through the pig colon (see Fig. \ref{Fig_experiment_3MA}(b-c)). This is because the rotating actuation can make the intestine stretch open during the capsule movement, which inherently reduces the friction to allow easier advancement of the capsule. It can also be observed in Fig. \ref{Fig_experiment_3MA}(b) that when CRMA is applied for capsule propulsion, the capsule is sometimes deflected to the right (see the snapshots at 24s and 54s) because of the continuous rotation in a single direction. In contrast, the heading direction of the capsule is better maintained during propulsion when using the RRMA method, as shown in Fig. \ref{Fig_experiment_3MA}(c).
The results demonstrate that the RRMA method can effectively reduce the environmental resistance and achieve better performance in the automatic propulsion of the capsule, which has the potential to realize safe and efficient inspection of the intestine.

Since the length of the human small intestine is about $6m$ \cite{hounnou2002anatomical}, assuming that the moving speed of the capsule in the human intestine is the same as that in the pig colon, it can be roughly estimated that it would take about $40$ minutes to traverse the small intestine with active WCE using our RRMA-based SMAL system. Therefore, this technology holds great promise to shorten the examination time in the small intestine compared with current passive WCE ($\sim247.2$ min) \cite{liao2010fields}\cite{ge2007clinical}, thereby improving the clinical acceptance of this non-invasive and painless screening technique.

\section{Conclusions}

In this paper, we have presented a novel actuation approach for active WCE named the \textit{reciprocally rotating magnetic actuation} (RRMA), to replace the conventional \textit{continuously rotating magnetic actuation} (CRMA), in order to mitigate the risk of causing intestinal malrotation and volvulus during actuation to improve patient safety. Moreover, the reciprocal rotation of the capsule under RRMA can make the intestine stretch open and reduce the environmental resistance, thereby improving the actuation efficiency in the narrow tubular environment. Based on the proposed RRMA method, we have developed a SMAL workflow to achieve automatic propulsion of a capsule in unknown tubular environments. The performance of our method is validated in real-world experiments conducted in an ex-vivo pig colon. Some conclusions are drawn as follows:

\begin{enumerate}
\item Compared with CRMA, the use of RRMA for rotating magnetic actuation of a capsule can effectively reduce the risk of causing intestinal malrotation and volvulus and enhance patient safety.
\item The proposed RRMA method can effectively reduce the environmental resistance compared with dragging-based actuation methods, and has the potential to realize more efficient and robust propulsion of the capsule in the intestinal environment.
\item The overall RRMA-based SMAL system can effectively achieve closed-loop actuation and localization of a magnetic capsule to automatically propel it in unknown tubular environments.
\end{enumerate}

By allowing the automatic propulsion of a capsule in an unknown tubular environment, the presented magnetic actuation method for active WCE has the potential to improve patient safety, shorten the examination time and improve the diagnostic accuracy and efficiency. Some other methods have also been presented to reduce tissue deformation in rotating magnetic actuation systems by reducing the attractive magnetic force with a specific motion of the actuator \cite{mahoney2013managing}\cite{mahoney2011managing}. Readers can draw inspiration from these different approaches to design more intelligent systems and methods to enhance patient safety in the magnetic manipulation of a capsule robot.
Also, this work only uses the feedback from magnetic sensors, and other sensors such as a camera may be included in the future studies to further improve the system performance. For instance, a vision-based method \cite{turan2018unsupervised}\cite{bao2014video} may be used to find the direction where the lumen is stretching open, which can better improve the propulsion efficiency. 
It should also be noted that although our method uses the magnetic torque to realize the reciprocal rotation of the capsule, the capsule is actually pushed forward by the magnetic force (see Section~III-A) instead of the magnetic torque, which is different from the helical propulsion-based systems (e.g., \cite{mahoney2014generating}\cite{popek2017first}\cite{popek2016six}). The magnetic-force-based systems are known to have a limited scalability compared with the magnetic-torque-based ones when the distances become longer, because the magnitude of the magnetic force decreases much faster than the magnetic torque when the distance between the actuator and the capsule increases \cite{mahoney2013generating}. With the current system setup, the distance between the actuator and the capsule $d$ is set to $15cm$ in our experiments to ensure effective propulsion, which is not large enough for the application in a clinical setting. In view of a clinical translation, one possible solution is to use a permanent magnet with a larger volume or a higher magnetic flux density as the extracorporeal actuator to increase the magnetic force applied to the capsule at a longer distance, thereby improving the applicability of the system on real patients.

Despite the challenges that need to be tackled toward clinical translation, the method presented in this paper has provided a promising solution to safely and effectively actuate a capsule in an unknown tubular environment, and the basic ideas proposed in this paper can hopefully be applied to the design of other medical robots working in tubular environments.


%

\ifCLASSOPTIONcaptionsoff
  \newpage
\fi



%

\bibliographystyle{IEEEtran}
\bibliography{bare_jrnl}




%

\begin{IEEEbiography}[{\includegraphics[width=1in,height=1.25in,clip,keepaspectratio]{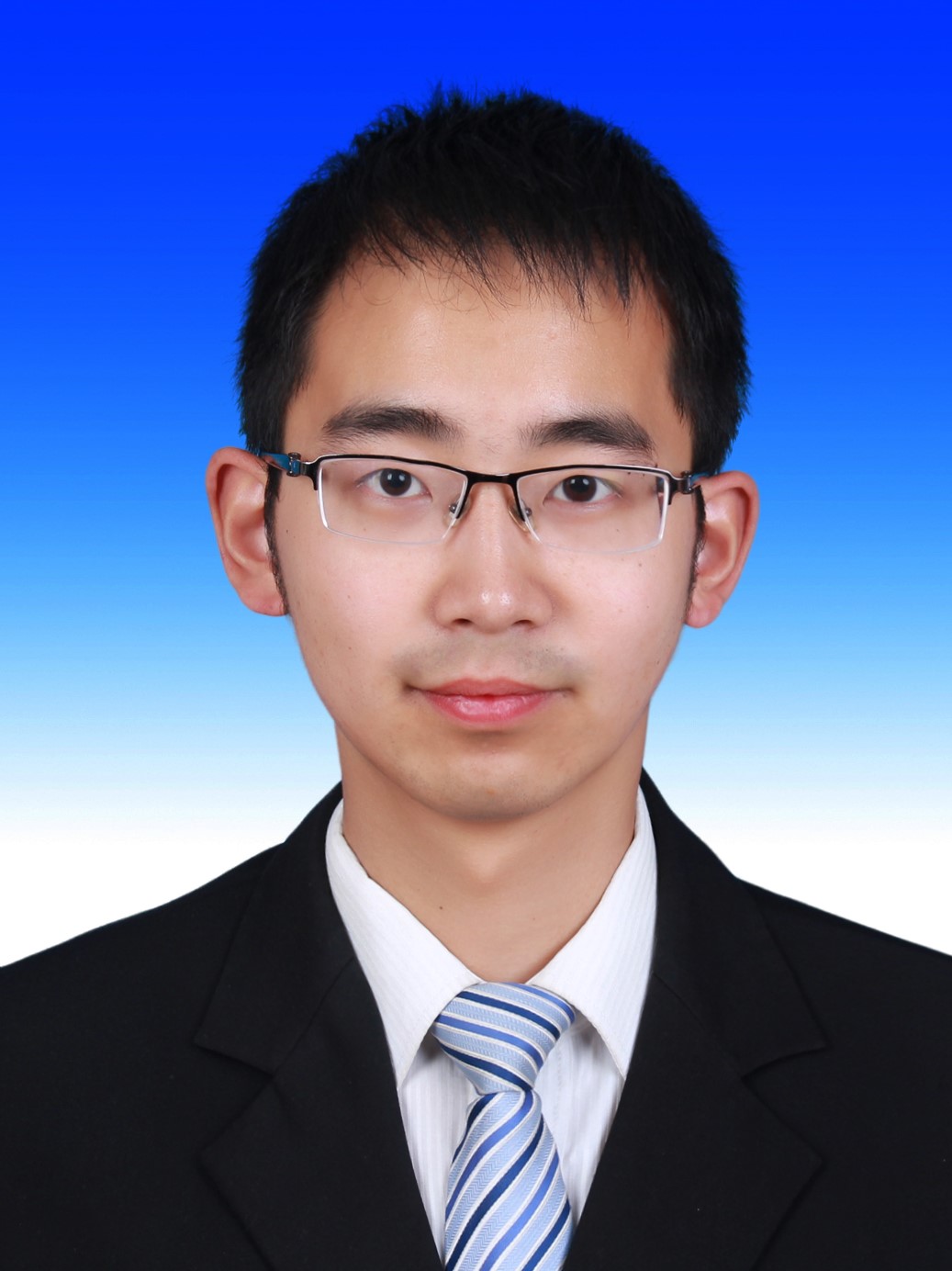}}]{Yangxin Xu}

received the B.Eng. degree in electrical engineering and its automation from Harbin Institute of Technology at Weihai (HIT), Weihai, China, in 2017. He is currently pursuing the Ph.D. degree with the Department of Electronic Engineering, The Chinese University of Hong Kong (CUHK), Hong Kong SAR, China.

His research focuses on magnetic actuation and localization methods and hardware implementation for active wireless robotic capsule endoscopy, supervised by Prof. Max Q.-H, Meng.

Mr. Xu received the Best Conference Paper from the 2018 IEEE International Conference on Robotics and Biomimetics (ROBIO), Kuala Lumpur, Malaysiain, in 2018.

\end{IEEEbiography}
\vfill

\begin{IEEEbiography}[{\includegraphics[width=1in,height=1.25in,clip,keepaspectratio]{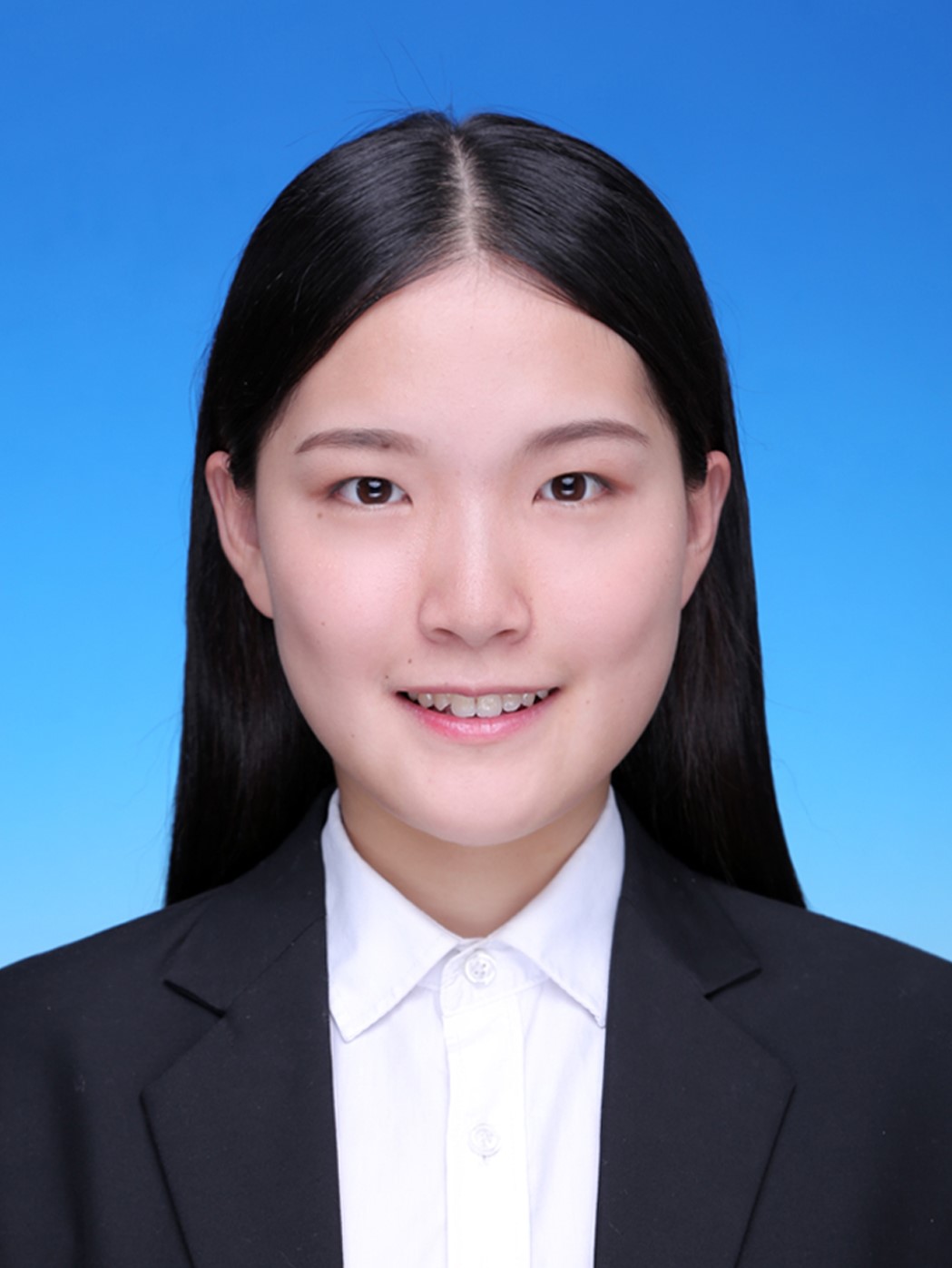}}]
{Keyu Li} received the B.Eng. degree in communication engineering from Harbin Institute of Technology at Weihai (HIT), Weihai, China, in 2019. She is currently pursuing the Ph.D. degree with the Department of Electronic Engineering, The Chinese University of Hong Kong (CUHK), Hong Kong SAR, China.

Her research focuses on artificial intelligence and medical robotic systems, supervised by Prof. Max Q.-H, Meng.
\end{IEEEbiography}
\vfill

\begin{IEEEbiography}[{\includegraphics[width=1in,height=1.25in,clip,keepaspectratio]{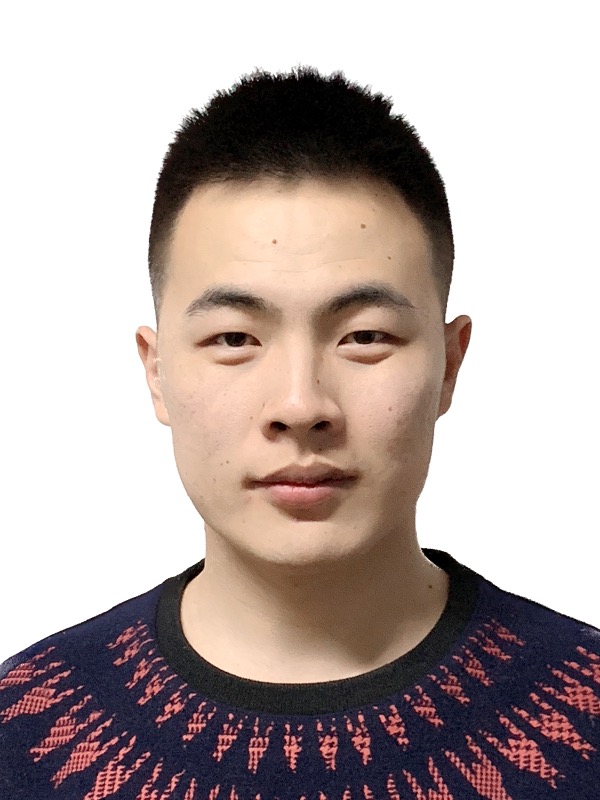}}]
{Ziqi Zhao} recived the B.Eng. and M.Eng. degree in mechanical engineering from Shenyang Jianzhu University, Shenyang, China, in 2016 and 2019, respectively. He is currently pursuing the Ph.D. degree with the Department of Electronic and Electrical Engineering, the Southern University of Science and Technology (SUSTech), Shenzhen, China.

His current research interests include bionic, medical and service robotics, supervised by Prof. Max Q.-H, Meng.

\end{IEEEbiography}
\vfill

\begin{IEEEbiography}[{\includegraphics[width=1in,height=1.25in,clip,keepaspectratio]{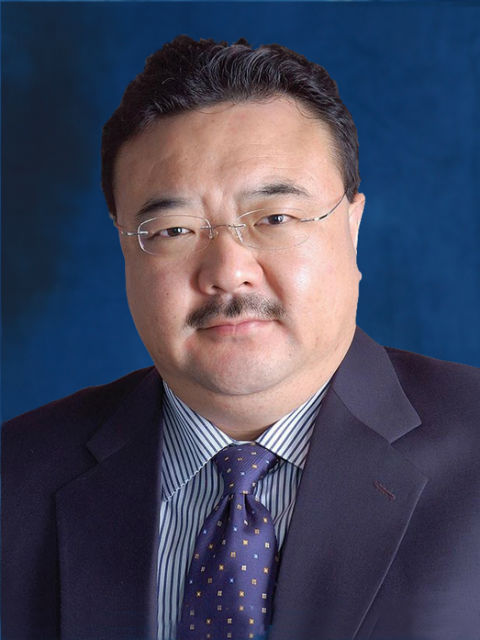}}]
{Max Q.-H. Meng} received the Ph.D. degree in electrical and computer engineering from the University of Victoria, Victoria, BC, Canada, in 1992.

He was with the Department of Electrical and Computer Engineering, University of Alberta, Edmonton, AB, Canada, where he served as the Director of the Advanced Robotics and Teleoperation Laboratory, holding the positions of Assistant Professor, Associate Professor, and Professor in 1994, 1998, and 2000, respectively. In 2001, he joined The Chinese University of Hong Kong, where he served as the Chairman of the Department of Electronic Engineering, holding the position of Professor. He is affiliated with the State Key Laboratory of Robotics and Systems, Harbin Institute of Technology, and is the Honorary Dean of the School of Control Science and Engineering, Shandong University, China. He is currently with the Department of Electronic and Electrical Engineering, Southern University of Science and Technology, on leave from the Department of Electronic Engineering, The Chinese University of Hong Kong, Hong Kong SAR, China, and also with the Shenzhen Research Institute of the Chinese University of Hong Kong, Shenzhen, China. His research interests include robotics, medical robotics and devices, perception, and scenario intelligence. He has published about 600 journal and conference papers and led more than 50 funded research projects to completion as PI.

Dr. Meng is an elected member of the Administrative Committee (AdCom) of the IEEE Robotics and Automation Society. He is a recipient of the IEEE Millennium Medal, a fellow of the Canadian Academy of Engineering, and a fellow of HKIE. He has served as an editor for several journals and also as the General and Program Chair for many conferences, including the General Chair of IROS 2005 and the General Chair of ICRA 2021 to be held in Xi'an, China.
\end{IEEEbiography}
\vfill




\end{document}